\documentclass[11pt]{article}
\usepackage[margin=1in]{geometry}
\usepackage{times}
\usepackage{placeins}
\usepackage{float}
\usepackage{microtype}
\usepackage{amsmath,amssymb,amsfonts}
\usepackage{graphicx}
\usepackage{algorithm}
\usepackage{algpseudocode}
\usepackage{hyperref}
\usepackage{iclr2026_conference}
\iclrfinalcopy
\usepackage{booktabs}
\usepackage{enumitem}
\usepackage{subcaption}
\usepackage{multirow}
\usepackage{pgfplots}
\usepackage{pgfplotstable}
\usepackage{tikz}
\pgfplotsset{compat=1.18}

\title{Recurrent Reasoning on Symbolic Puzzles with Sequence Models}

\author{
Gowrav Mannem$^{1}$, Chowdhury Marzia Mahjabin$^{1}$, Jason Chen$^{1,2}$, Shivank Garg$^{1}$, Kevin Zhu$^{1}$ \\[4pt]
\small $^{1}$ Algoverse AI Research\quad $^{2}$ Cornell University \\
\small \texttt{shivank@algoverseairesearch.org}
}

\begin{document}
\maketitle
\begin{abstract}
Large language models often appear strong on symbolic and algorithmic tasks,
yet this apparent strength can hide brittle behaviour when problems become
longer, harder, or slightly out of distribution. A major limitation of
current reasoning benchmarks is that many primarily test whether a model
can produce a valid answer, while paying less attention to whether the
solution is minimal, robust, and stable under controlled difficulty scaling.
We introduce RecurrReason, a difficulty-controlled benchmark of
four recurrent logic puzzles (Tower of Hanoi, River Crossing, Block World,
and Checkers Jumping) with BFS-optimal trajectories and a single
interpretable difficulty parameter $N \in \{1,\dots,10\}$, totalling
10{,}817 unique puzzles and 285{,}933 moves.
We benchmark two Transformer families, an encoder-decoder model
(T5-style) and a decoder-only model (GPT-2-style), under consistent data
splits and evaluation criteria, training on $N{=}1$ to $7$ and evaluating on
both held-out in-distribution instances and harder out-of-distribution
instances at $N{=}8$ to $10$.
Fine-tuned pre-trained T5 achieves 97.27\% validation and 81.00\% OOD
accuracy on Block World; all models score 0.00\% on River Crossing under all conditions.
Failure mode analysis reveals that architecture is a stronger determinant
of success than scale. Pre-training transfers only to puzzles with locally
structured transition functions. Our code and dataset will be open-sourced upon acceptance. 
\end{abstract}
\section{Introduction}

Recent progress in neural language models has made multi-step reasoning look
increasingly accessible. Models can generate coherent intermediate steps,
imitate solver-like behaviour, and often reach the correct final answer on
curated reasoning tasks. However, for algorithmic problems, correctness is not
only about producing a plausible endpoint. A model must repeatedly choose valid
intermediate actions, avoid illegal transitions, preserve consistency across
long horizons, and ideally reach the goal with a minimal-length solution.
These requirements become more demanding as problem size grows, making
algorithmic reasoning a useful setting for separating genuine robustness from
shallow pattern matching.

Recent work~\citep{shojaee2025illusion} argues that models can look competent while relying on brittle heuristics that collapse under modest difficulty changes. We study this concern in a setting where every intermediate step is mechanically verifiable.
Recurrent logic puzzles are ideal for this: they are fully state-based with explicit transition rules, allowing us to separately evaluate move validity, goal attainment, and near-optimal efficiency.
A discussion of related work is in Appendix~\ref{app:related_work}. Our contributions include:

\begin{itemize}[leftmargin=1.5em,itemsep=0.25em]
  \item We introduce a unified benchmark (RecurrReason) of four recurrent logic puzzles with BFS-optimal solutions, scalable difficulty, and stepwise supervision (10{,}817 puzzles, 285{,}933 moves).
  \item We provide a controlled comparison of two Transformer families (T5-style encoder-decoder and GPT-2-style decoder-only) under consistent preprocessing and splits.
  \item We design an evaluation protocol focused on complete trajectory rollouts and reasoning-specific metrics (syntax validity, move validity, termination accuracy, and optimality gap).
  \item We show that puzzle structure (specifically transition locality, action space size, and solution length growth) is the primary determinant of learnability.
\end{itemize}

\section{RecurrReason Benchmark}

A recurrent reasoning game is defined by: (1)~a finite state set $S$, (2)~actions $A$ with constraint-respecting transitions, (3)~a goal set $G \subseteq S$, (4)~recurrent multi-step structure, and (5)~constraint satisfaction requiring systematic search. We extend four puzzles from \citet{shojaee2025illusion} with permutation augmentations, BFS-optimal trajectories, and autoregressive formats, yielding \textit{10{,}817 puzzles} and \textit{285{,}933 moves} (Appendix~\ref{app:dataset}).

\begin{table}[ht]
  \caption{RecurrReason benchmark statistics summary.
           Full table in Appendix~\ref{app:dataset}.}
  \label{tab:puzzle_stats}
  \centering
  \resizebox{\textwidth}{!}{%
  \begin{tabular}{l rr rr rr rr}
    \toprule
    & \multicolumn{2}{c}{\textbf{Block World}}
    & \multicolumn{2}{c}{\textbf{Checkers Jumping}}
    & \multicolumn{2}{c}{\textbf{Tower of Hanoi}}
    & \multicolumn{2}{c}{\textbf{River Crossing}} \\
    \cmidrule(lr){2-3}\cmidrule(lr){4-5}\cmidrule(lr){6-7}\cmidrule(lr){8-9}
    $N$ & \# Puzz. & Moves & \# Puzz. & Moves
        & \# Puzz. & Moves & \# Puzz. & Moves \\
    \midrule
    1 to 7  & 549 & 3{,}214 & 2{,}700 & 76{,}612  & 42 & 1{,}482 & 630  & 2{,}230 \\
    8 to 10 & 300 & 2{,}613 & 3{,}000 & 165{,}882 & 18 & 10{,}734 & 3{,}578 & 23{,}166 \\
    \midrule
    \textbf{Total} & 849 & 5{,}827 & 5{,}700 & 242{,}494
                   & 60 & 12{,}216 & 4{,}208 & 25{,}396 \\
    \bottomrule
  \end{tabular}%
  }
\end{table}

\section{Task Formulation, Training, and Evaluation}
\label{sec:method_eval}

Each puzzle is cast as a predict-the-next-step task. States are serialized as strings (e.g.\ disk-to-peg lists for ToH, stack contents for BW). From each BFS-optimal trajectory $(s_0,\dots,s_T)$ we create step pairs $(s_t \to s_{t+1})$. At evaluation, the model rolls out autoregressively: $\hat{s}_{t+1} = f_\theta(\hat{s}_t, g),\; \hat{s}_0 = s_0$, with no ground-truth states provided.

\paragraph{Models.}
\label{sec:models}
Both architectures learn $f_\theta:(s_t,g)\mapsto \hat{s}_{t+1}$.
\textbf{T5-small} (60M)~\citep{raffel2020exploring}: the encoder computes bidirectional attention over $[s_t; s_g]$, so every token in $s_t$ attends to every token in $s_g$ before decoding begins; the decoder generates $\hat{s}_{t+1}$ via cross-attention to this full representation, making the goal a first-class conditioning signal at every step.
\label{sec:t5}
\textbf{GPT-2} (124M)~\citep{radford2019language}: the causal mask prevents $s_t$ tokens from attending to $s_g$ (which appears later in the concatenated sequence $[s_t; s_g; \hat{s}_{t+1}]$), creating a structural bottleneck for goal-directed planning.
\label{sec:gpt2}
Full architecture equations are in Appendix~\ref{app:arch_details}. Notation is defined in Appendix~\ref{app:notation}.

\paragraph{Training conditions.} We evaluate three experimental conditions for each architecture: (1) trained from scratch on puzzle data only, (2) pre-trained zero-shot (ZS), where we evaluate the base pre-trained checkpoint without any puzzle-specific training, and (3) pre-trained fine-tuned (FT),  where we initialize from pre-trained checkpoints and fine-tune on puzzle data.

\paragraph{Data and splits.}
\label{sec:data}
We enumerate instances per $N$, compute BFS-optimal trajectories, apply puzzle- specific augmentations, and serialize into model-ready inputs (Appendix E). We split instances  from N=1 to 7 into training and validation sets (80/20 random split within each N); N=8 to 10 
serves as out-of-distribution (OOD) evaluation. Results reported as "Val (\%)" refer to held-out validation instances from N=1-7; "OOD (\%)" refers to all instances from N=8-10.

\paragraph{Training.}
All models minimise token-level cross-entropy on predicted next states, masking \texttt{<PAD>} positions. Both use AdamW~\citep{loshchilov2019decoupled} ($\text{LR}=10^{-4}$, batch 16) with early stopping. Full training hyperparameters are shown in Appendix~\ref{app:hparams}.

\paragraph{Evaluation and metrics.}
Rollouts terminate at goal reached, unparseable output, illegal transition, or horizon $T_{\max}$ exceeded. Primary metrics are trajectory success rate, move legality rate, and optimality gap $= (|\hat{\tau}| - |\tau^\star|)/|\tau^\star|$, (computed only for correctly solved puzzles), which measures the percentage of excess steps relative to the BFS-optimal solution length.

\section{Results}
\label{sec:results}

A puzzle is solved only if the rollout reaches $s_g$ via valid states within $T_{\max}=2|\tau^\star|$ steps (notation in Appendix~\ref{app:notation}). Each failed rollout is classified into one of four modes: \textbf{invalid\_move} (constraint violation), \textbf{invalid\_output} (unparseable), \textbf{loop} (cycle), or \textbf{premature\_stop} (halts before goal).

\subsection{Cross-Puzzle Summary}

Table~\ref{tab:cross_puzzle} summarises the best result per puzzle. Block World is the only puzzle with substantial learning; the other three yield at most 1.11\% validation and 0.10\% OOD, establishing that puzzle structure determines learnability more than architecture or pre-training. Checkers Jumping shows identical performance for both T5 and GPT-2 because both models only solve trivial instances where the start state equals the goal state, failing on all instances requiring actual moves.

\begin{table}[ht]
  \caption{Cross-puzzle summary: best model result per puzzle.
           Full per-puzzle result tables are in Appendix~\ref{app:full_tables}.}
  \label{tab:cross_puzzle}
  \centering
  \begin{tabular}{lllcc}
    \toprule
    \textbf{Puzzle} & \textbf{Best Model} & \textbf{Cond.}
                    & \textbf{Val (\%)} & \textbf{OOD (\%)} \\
    \midrule
    Block World      & T5 (pre-trained) & Fine-tuned & \textbf{97.27} & \textbf{81.00} \\
    Tower of Hanoi   & T5 (pre-trained) & Fine-tuned & 11.11          & 0.00 \\
    Checkers Jumping & T5/GPT-2 (PT)    & Fine-tuned & 1.11           & 0.10 \\
    River Crossing   & \textit{none}    & n/a        & 0.00           & 0.00 \\
    \bottomrule
  \end{tabular}
\end{table}

Block World succeeds where others fail for three compounding reasons.
First, its transition function is \emph{local}: whether a block can be moved requires checking only the top of its source stack ($O(1)$ tokens), so the model does not need to reason over the entire board to produce a valid move.
Second, its solution lengths grow linearly ($L(N)=O(N)$), so compounding rollout errors accumulate slowly.
Third, the training signal is dense and consistent: each of the 549 training puzzles shares the same move grammar, giving the model many opportunities to generalise the same rule.
The other three puzzles violate at least one of these conditions: ToH has exponential solution length ($2^N-1$), RC requires global constraint verification at every step ($O(N)$ tokens), and CJ combines quadratic solution length with a constrained jump grammar that appears only in very few valid configurations.

\paragraph{Zero-result puzzles.}
ToH: only 1/9 validation puzzles solved ($N{=}1$, one move); all fail at $N{\geq}2$ because $L(N)=2^N-1$ requires recursive decomposition a flat mapping cannot represent.
CJ: 1.11\% validation corresponds exclusively to trivial instances where start equals goal.
RC: 0.00\% everywhere despite low training loss; the global safety constraint and combinatorial action space (up to 175 candidates at $N{=}5$) defeat all models.
Full tables: Appendix~\ref{app:full_tables}.

T5 pre-trained FT achieves 100\% at $N{=}1$ to $2$, stays above 93\% through $N{=}7$, and degrades gradually to 75\% at $N{=}10$ (Figure~\ref{fig:bw_by_n} in Appendix~\ref{app:bw_figure}), consistent with rule generalisation rather than memorisation.
The gradual OOD degradation (84\% at $N{=}8$, 84\% at $N{=}9$, 75\% at $N{=}10$) contrasts with the abrupt collapse seen in GPT-2, which reaches 21 to 25\% validation but drops to 0\% OOD across all conditions.
GPT-2's failure pattern shifts systematically: at validation, loops account for 80 to 92\% of failures, while at OOD, invalid\_move dominates ($>$91\%).
This transition indicates that GPT-2 learns which moves are \emph{legal} within the training distribution but cannot select among them based on goal proximity, a direct consequence of the causal attention bottleneck described in Section~\ref{sec:method_eval}.

Failure mode breakdowns are visualised in Figure~\ref{fig:fm_all} (Appendix~\ref{app:fm_figure}) and tabulated in Appendix~\ref{app:failure_tables}.

\section{Discussion}
\label{sec:analysis}

\paragraph{Architecture matters more than scale.}
T5 (60M) outperforms GPT-2 (124M) on every puzzle despite having half the parameters, because its encoder provides full bidirectional attention over $[s_t; s_g]$, making the goal a first-class conditioning signal at every decoder step. GPT-2's causal mask prevents $s_t$ from attending to $s_g$, creating a structural bottleneck for goal-directed planning. \citet{ding2024causallm} proved theoretically that causal language models converge to a suboptimal solution compared to prefix (bidirectional) models, formalising why our goal tokens cannot be optimally utilised under a causal mask. Empirically, \citet{wang2022language} showed that encoder-decoder models with non-causal attention outperform decoder-only models of comparable size after multitask fine-tuning, and \citet{zhang2022examining} found that architectural differences have the largest impact at small scales, exactly the regime of our experiments. \citet{csordas2021devil} and \citet{tay2023transcending} further demonstrated that targeted architectural modifications improve systematic generalization and can transcend scaling laws, consistent with our finding that a well-matched architecture at 60M parameters surpasses a mismatched one at 124M.

\paragraph{Pre-training benefit is puzzle-dependent.}
All models score 0.00\% zero-shot. After fine-tuning, pre-training helps only where the transition function is local: T5 gains +97.27 pp on Block World ($O(1)$ verification) but +0.00 pp on River Crossing ($O(N)$ global constraint). GPT-2 pre-training adds only 2.73 pp on BW, confirming the architectural bottleneck limits what pre-training can contribute. \citet{talmor2020olmpics} showed empirically that pre-training fails on half of their symbolic reasoning tasks, with gains appearing only where task structure overlaps with natural language distributional patterns. \citet{mueller2022coloring} explained the mechanism: pre-training imparts a hierarchical inductive bias that helps only on tasks with locally decomposable structure. \citet{papadimitriou2020learning} demonstrated that transfer depends on shared structural properties between pre-training and target domains, even across modalities. \citet{furrer2020compositional} confirmed this on compositional generalization benchmarks, finding that pre-training helps on locally decomposable splits but fails on globally compositional ones, a precise parallel to our Block World vs.\ River Crossing results.

\paragraph{Failure modes.}
Invalid\_move reflects constraint non-generalisation~\citep{mccoy2019right}. Loops on BW signal partial rule learning (legal moves but no goal direction). Premature stopping in GPT-2 on ToH results from the statistical overrepresentation of $\langle\texttt{STOP}\rangle$ in low-diversity training data. Full breakdowns: Appendix~\ref{app:failure_tables}.

\paragraph{Learnability is determined by puzzle structure.}
Three properties set the ceiling: (1) \emph{transition locality} ($O(1)$ for BW vs.\ $O(N)$ for RC), (2) \emph{action space size} ($O(K^2)$ for BW vs.\ combinatorial for RC), and (3) \emph{solution length growth}. Assuming an independent per-step error rate $\epsilon$ (small), success probability compounds across the solution trajectory:
\begin{equation}
  P(\text{success}) = (1 - \varepsilon)^{L(N)} \approx e^{-\varepsilon L(N)}.
  \label{eq:compounding}
\end{equation}
BW has $L(N)=O(N)$; CJ has $(N{+}1)^2{-}1$; ToH has $2^N{-}1$. This compounding effect is a direct consequence of autoregressive rollout under imitation learning~\citep{ross2011reduction}. This explains why even T5's $N{=}1$ ToH success does not extend to $N{\geq}2$. These findings align with \citet{valmeekam2022large} and \citet{kambhampati2024can}: reliable planning requires search augmentation beyond greedy next-step prediction. Full derivations of solution length formulas, transition locality, action space sizes, and the compounding error model are in Appendix~\ref{app:learnability}.


\section{Conclusion}

We introduce RecurrReason, a difficulty-controlled benchmark of four recurrent logic puzzles with BFS-optimal solutions and systematic failure mode analysis. Our comparison of encoder-decoder (T5)  and decoder-only (GPT-2) Transformers reveals that architectural inductive biases matter more than 
scale: bidirectional goal-conditioning enables T5 to achieve 81\% OOD accuracy on Block World, while GPT-2's causal attention bottleneck prevents effective goal-directed planning across all puzzles. Crucially, pre-training transfers only to puzzles with local transition functions. Block World succeeds due to $O(1)$ transition verification and linear solution growth, while River Crossing's global $O(N)$ constraints and Tower of Hanoi's exponential solution length $(2^N-1)$ defeat all models despite low training loss. This structural determinism, formalized through our compounding error model, suggests that reliable multi-step reasoning requires either architectural search capabilities or explicit symbolic constraint checking, not merely larger language models. Our benchmark and evaluation protocol provides a controlled testbed for future work on length generalization, goal-conditioned planning, and the architectural requirements for systematic 
algorithmic reasoning.




\bibliographystyle{plainnat}
\bibliography{iclr2026_conference}

\appendix

\section{Related Work}
\label{app:related_work}

\paragraph{Neural algorithmic reasoning.}
Teaching neural networks to execute algorithms has a long history, from
LSTMs trained on simple programs~\citep{zaremba2015learning} and Neural
Turing Machines with external memory~\citep{graves2014neural} to the
compositional Neural Programmer-Interpreter~\citep{reed2016neural}.
More recently, graph neural networks have been applied to classical
algorithms via the CLRS benchmark~\citep{velickovic2022clrs}, and
\citet{velickovic2020neural} showed that GNNs
can learn to execute graph algorithms by imitating intermediate execution
traces.
Our work differs in that we study whether plain Transformer
sequence models, without graph structure or memory augmentation, can
learn recurrent symbolic puzzles purely from textual state-action traces.

\paragraph{Length generalization in Transformers.}
Transformers notoriously struggle to generalize to input lengths unseen
during training.
\citet{anil2022exploring} demonstrated that scratchpad
prompting can help, while \citet{zhou2024algorithms}
proposed the RASP-Generalization conjecture linking length
generalization to the existence of short RASP-L programs.
\citet{lee2024teaching} showed that small Transformers can
learn arithmetic with appropriate data formatting and curriculum.
\citet{deletang2023neural} connected architectural
expressiveness to the Chomsky hierarchy, finding that vanilla
Transformers fail on tasks beyond regular languages.
Our OOD evaluation ($N{=}8$ to $10$ after training on $N{=}1$ to $7$) is a
direct test of length generalization; we find that success depends
heavily on puzzle structure rather than model scale.

\paragraph{Intermediate computation and chain of thought.}
\citet{nye2021show} introduced scratchpads (supervised
intermediate steps) that dramatically improve multi-step reasoning.
\citet{wei2022chain} scaled this idea to large models via
chain-of-thought prompting.
RecurrReason adopts a similar philosophy: models are trained on full
step-by-step solution trajectories, giving them access to every
intermediate state, yet this supervision alone does not guarantee
generalization.

\paragraph{Planning with neural networks.}
Classical planning remains difficult for neural models.
\citet{valmeekam2023planbench} showed with PlanBench
that LLMs perform poorly on standard planning domains.
\citet{lehnert2024beyond} trained Transformers to
imitate A* search dynamics and achieved strong planning performance,
but required explicit search traces.
The Decision Transformer~\citep{chen2021decision} cast reinforcement
learning as goal-conditioned sequence modeling, conditioning on desired
returns to generate actions.
Our setup is related: the model receives a goal state and must produce
an optimal action sequence, but we operate on fully deterministic
symbolic puzzles with verifiable BFS-optimal solutions rather than
stochastic MDPs.

\paragraph{Compositional generalization and reasoning benchmarks.}
\citet{lake2018generalization} introduced SCAN to expose
systematic generalization failures in sequence models.
\citet{dziri2023faith} showed that Transformers reduce
multi-step compositional reasoning to linearized subgraph matching,
with accuracy decaying as task complexity grows.
\citet{saxton2019analysing} benchmarked neural models on
mathematics problems of varying difficulty.
RecurrReason complements these benchmarks by providing a setting where
difficulty is controlled by a single integer parameter $N$, every
intermediate step is mechanically verifiable, and the recursive
structure of the puzzles exposes whether models learn genuine
algorithmic strategies or merely memorize shallow patterns.

\section{Architecture Details}
\label{app:arch_details}

\subsection{Encoder-Decoder Transformer (T5)}

T5-small~\citep{raffel2020exploring} is a 60M-parameter encoder-decoder
Transformer pre-trained on C4. Given the concatenation $x = [s_t;\, s_g]$,
the encoder computes a bidirectional contextual representation:
\begin{equation}
  H = \mathrm{Enc}(x) \in \mathbb{R}^{|x| \times d},
  \qquad
  A_{ij} \propto \exp\!\left(\frac{\mathbf{q}_i^\top \mathbf{k}_j}{\sqrt{d}}\right),
  \label{eq:t5enc}
\end{equation}
where $\mathbf{q}_i, \mathbf{k}_j \in \mathbb{R}^d$ are query and key vectors.
The decoder generates $\hat{s}_{t+1}$ token-by-token via cross-attention:
\begin{equation}
  P\!\left(\hat{s}_{t+1}^{(\ell)} \mid \hat{s}_{t+1}^{(1:\ell-1)}, H\right)
    = \mathrm{softmax}\!\left(W_O\,
      \mathrm{CrossAttn}\!\left(\hat{s}_{t+1}^{(1:\ell-1)},\, H\right)\right).
  \label{eq:t5dec}
\end{equation}

\subsection{Decoder-Only Transformer (GPT-2)}

GPT-2~\citep{radford2019language} (124M parameters) concatenates state and goal
into a flat causal sequence $[s_t;\, s_g;\, \hat{s}_{t+1}]$:
\begin{equation}
  p_\theta(\hat{s}_{t+1}) = \prod_{\ell=1}^{L}
    p_\theta\!\left(\hat{s}_{t+1}^{(\ell)} \mid \hat{s}_{t+1}^{(1:\ell-1)},
      s_t, s_g\right).
  \label{eq:gpt2}
\end{equation}
Training minimises $\mathcal{L}(\theta) = -\sum_{\ell} \log p_\theta(\hat{s}_{t+1}^{(\ell)} \mid \cdot)$.
The causal mask enforces $M_{ij} = -\infty$ for $j > i$, preventing $s_t$ from
attending to $s_g$.

\section{Block World Per-$N$ Accuracy}
\label{app:bw_figure}

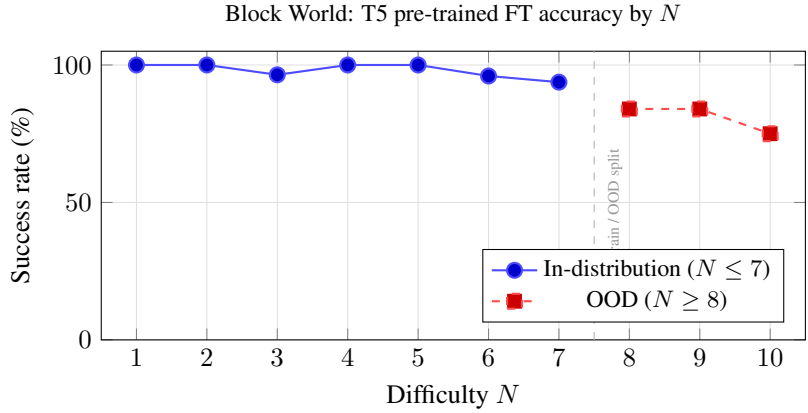
\begin{figure}[H]
  \centering
  \begin{tikzpicture}
    \begin{axis}[
      width=0.78\textwidth, height=5.4cm,
      xlabel={Difficulty $N$}, ylabel={Success rate (\%)},
      ymin=0, ymax=105, xmin=0.5, xmax=10.5,
      xtick={1,...,10},
      grid=both, grid style={line width=0.3pt, draw=gray!25},
      legend style={at={(0.97,0.06)}, anchor=south east, font=\small},
      title={Block World: T5 pre-trained FT accuracy by $N$},
      title style={font=\small}, mark size=2.8pt,
    ]
      \addplot+[color=blue!70, mark=*, thick] coordinates {
        (1,100)(2,100)(3,96.43)(4,100)(5,100)(6,96.00)(7,93.75)};
      \addplot+[color=red!70, mark=square*, thick, dashed] coordinates {
        (8,84)(9,84)(10,75)};
      \draw[dashed, gray!60, thin] (axis cs:7.5,0) -- (axis cs:7.5,105);
      \node[font=\tiny, gray!80, rotate=90] at (axis cs:7.8,50)
        {train / OOD split};
      \legend{In-distribution ($N \leq 7$), OOD ($N \geq 8$)}
    \end{axis}
  \end{tikzpicture}
  \caption{T5 pre-trained FT on Block World per $N$.
           Performance is high throughout training and degrades gradually at OOD, not abruptly.}
    \label{fig:bw_by_n}
\end{figure}

Figure~\ref{fig:bw_by_n} shows the per-difficulty accuracy of T5 pre-trained FT on Block World. Performance remains high across in-distribution levels and degrades gradually at OOD, consistent with genuine rule generalisation rather than memorisation.

\section{Failure Mode Breakdown}
\label{app:fm_figure}

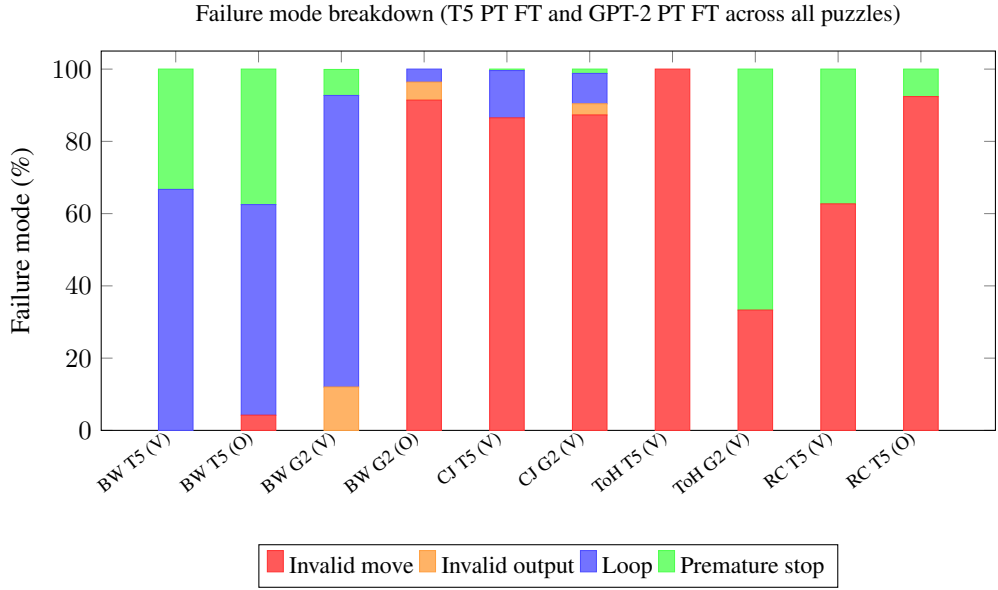
\begin{figure}[h]
  \centering
  \begin{tikzpicture}
    \begin{axis}[
      ybar stacked, bar width=13pt,
      width=0.96\textwidth, height=6.6cm,
      ymin=0, ymax=105, ylabel={Failure mode (\%)},
      xtick=data,
      symbolic x coords={
        BW T5 (V), BW T5 (O),
        BW G2 (V), BW G2 (O),
        CJ T5 (V), CJ G2 (V),
        ToH T5 (V), ToH G2 (V),
        RC T5 (V), RC T5 (O)},
      xticklabel style={rotate=38, anchor=east, font=\scriptsize},
      legend style={at={(0.5,-0.30)}, anchor=north, legend columns=4,
                    font=\small},
      title={Failure mode breakdown (T5 PT FT and GPT-2 PT FT across all puzzles)},
      title style={font=\small},
    ]
      \addplot+[fill=red!65,    draw=red!80]    coordinates {
        (BW T5 (V),0.0)  (BW T5 (O),4.2)
        (BW G2 (V),0.0)  (BW G2 (O),91.4)
        (CJ T5 (V),86.5) (CJ G2 (V),87.3)
        (ToH T5 (V),100) (ToH G2 (V),33.3)
        (RC T5 (V),62.7) (RC T5 (O),92.4)};
      \addplot+[fill=orange!60, draw=orange!75] coordinates {
        (BW T5 (V),0.0)  (BW T5 (O),0.0)
        (BW G2 (V),12.0) (BW G2 (O),5.0)
        (CJ T5 (V),0.0)  (CJ G2 (V),3.1)
        (ToH T5 (V),0.0) (ToH G2 (V),0.0)
        (RC T5 (V),0.0)  (RC T5 (O),0.0)};
      \addplot+[fill=blue!55,   draw=blue!70]   coordinates {
        (BW T5 (V),66.7) (BW T5 (O),58.3)
        (BW G2 (V),80.7) (BW G2 (O),3.6)
        (CJ T5 (V),13.1) (CJ G2 (V),8.4)
        (ToH T5 (V),0.0) (ToH G2 (V),0.0)
        (RC T5 (V),0.0)  (RC T5 (O),0.0)};
      \addplot+[fill=green!55,  draw=green!70]  coordinates {
        (BW T5 (V),33.3) (BW T5 (O),37.5)
        (BW G2 (V),7.2)  (BW G2 (O),0.0)
        (CJ T5 (V),0.4)  (CJ G2 (V),1.2)
        (ToH T5 (V),0.0) (ToH G2 (V),66.7)
        (RC T5 (V),37.3) (RC T5 (O),7.6)};
      \legend{Invalid move, Invalid output, Loop, Premature stop}
    \end{axis}
  \end{tikzpicture}
  \caption{Failure mode breakdown for T5 pre-trained fine-tuned (T5 PT FT) and GPT-2 pre-trained fine-tuned (G2 PT FT) across all four puzzles. V: validation set (held-out instances from N=1-7), 
O: OOD (N=8-10). BW: Block World, CJ: Checkers Jumping, ToH: Tower of Hanoi, RC: River Crossing. Values shown are percentages of failed rollouts only (solved rollouts excluded). Invalid move is a constraint violation. Invalid output is an unparseable state. Loop is a repeated state. Premature stop is when model halts before goal.}
  \label{fig:fm_all}
\end{figure}

Figure~\ref{fig:fm_all} presents the failure mode distribution across all four puzzles for the best-performing T5 and GPT-2 variants. Invalid moves dominate in globally constrained puzzles (ToH, RC), while loops are the primary failure mode for Block World, reflecting partial rule learning without goal-directed selection.

\section{Tower of Hanoi}\label{app:puzzles}

The Tower of Hanoi is a classic recursive puzzle consisting of three pegs (labeled A, B, and C) and N disks of different sizes, numbered from 1 (smallest) to N (largest). The puzzle is governed by three fundamental constraints. (1) Single Disk Movement: Only one disk may be moved at a time. (2) Top Disk Access: Only the topmost disk from any peg can be selected for movement. (3) Size Ordering Constraint: A larger disk may never be placed on top of a smaller disk. The objective is to transfer all disks from a designated start peg to a target end peg while maintaining size ordering (largest at bottom, smallest at top) throughout all intermediate states. The minimum number of moves required to solve the Tower of Hanoi with N disks is $2^N - 1$, making it an exponentially scaling problem that provides fine-grained control over computational complexity.

Our Tower of Hanoi dataset generation extends the approach described in \citet{shojaee2025illusion} with several key enhancements to create a more comprehensive and realistic evaluation benchmark. We implemented a recursive solution generator based on the classical Tower of Hanoi algorithm. This recursive function generates the optimal sequence of moves for transferring N disks from the start peg to the end peg using the auxiliary peg.

\begin{algorithm}[h]
\caption{Recursive Tower of Hanoi Sequence Generation}
\label{alg:hanoi}
\begin{algorithmic}[1]
\Procedure{Hanoi}{$n, \text{start}, \text{end}, \text{aux}, \text{moves}$}
    \If{$n = 1$}
        \State $\text{moves}.\text{append}([n, \text{start}, \text{end}])$
    \Else
        \State \Call{Hanoi}{$n - 1, \text{start}, \text{aux}, \text{end}, \text{moves}$}
        \State $\text{moves}.\text{append}([n, \text{start}, \text{end}])$
        \State \Call{Hanoi}{$n - 1, \text{aux}, \text{end}, \text{start}, \text{moves}$}
    \EndIf
\EndProcedure
\end{algorithmic}
\end{algorithm}

While the prior work primarily focused on the canonical configuration (start peg A, end peg C), we systematically generate all possible start-end peg combinations. This creates 6 distinct configurations for each problem size: peg A to peg B, peg A to peg C, peg B to peg A, peg B to peg C, peg C to peg A, peg C to peg B. This augmentation tests whether models can generalize the solution strategy across different spatial arrangements rather than potentially memorizing patterns for specific configurations.

Our dataset includes explicit state tracking at each step of the solution. Initial configuration with all disks on the starting peg (\texttt{start\_state}). Target configuration with all disks on the ending peg (\texttt{goal\_state}). State before applying each move (\texttt{current\_state}). State after applying each move (\texttt{next\_state}). Each state is represented as a list of three lists (one per peg), where each peg contains its disks in order from top to bottom. For example, \texttt{[[1, 2, 3], [], []]}\  represents disks one, two, and three stacked on peg A.

\begin{figure}[H]
    \centering
    \includegraphics[page=1, width=0.8\textwidth]{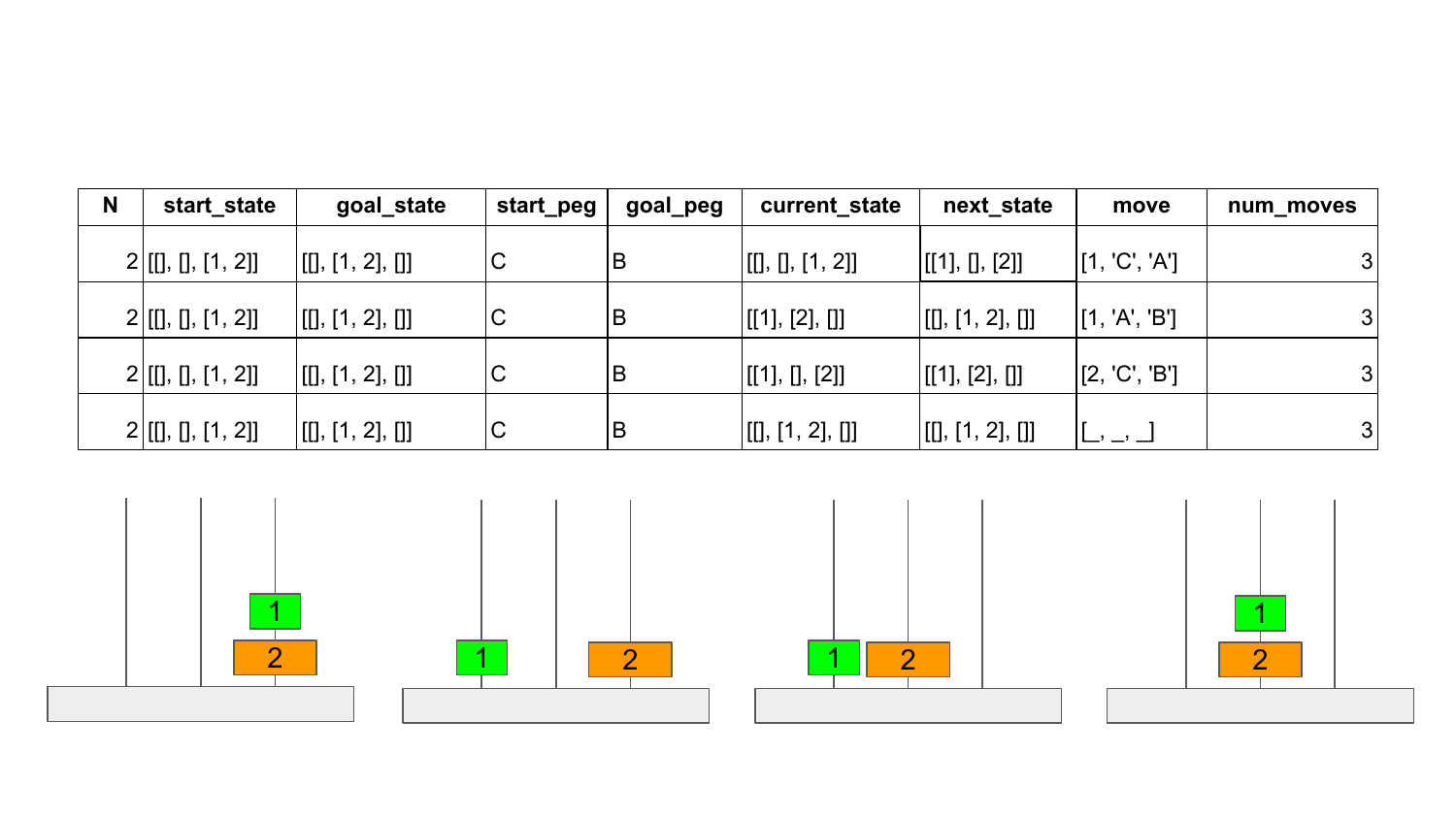}
    \caption{Example Tower of Hanoi puzzle with $N=3$ disks showing the initial state, goal state, and the optimal solution trajectory. \texttt{[\_,\_,\_]} to indicate puzzle completion.}
    \label{fig:puzzle_hanoi}
\end{figure}

A critical enhancement is our transformation to an expanded format where each row represents a single move within a solution trajectory. For a puzzle requiring M moves, we generate M+1 rows. Each row contains the current state, next state, and the move that connects them. The final row uses the sentinel move \texttt{[\_,\_,\_]} to indicate puzzle completion. This granular representation enables (1) auto-regressive training: Models can learn to predict the next optimal move given the current state and goal. (2) Intermediate verification: Each move can be validated independently using puzzle simulators. (3) Trajectory analysis: The complete solution path can be analyzed for consistency and optimality. Figure~\ref{fig:puzzle_hanoi} illustrates an example Tower of Hanoi configuration and its solution trajectory.

\section{Checkers Jumping}
Checkers Jumping is a one-dimensional constraint-satisfaction puzzle that tests sequential reasoning and planning capabilities. The puzzle consists of a linear arrangement of N red checkers (R), N blue checkers (B), and a single empty space \texttt{(\_)}, forming a board of length $2N + 1$. In the standard initial configuration, N red checkers are positioned on the left side, followed by an empty space in the middle, and N blue checkers on the right side. The objective is to swap the positions of all red and blue checkers, effectively mirroring the initial configuration to reach. The puzzle is governed by two fundamental movement rules: (1) Slide Movement: A checker can slide forward into an adjacent empty space. (2) Jump Movement: A checker can jump forward over exactly one checker of the opposite color to land in an empty space. (3) A critical constraint is that checkers cannot move backward toward their starting side. Red checkers can only move rightward, and blue checkers can only move leftward from the initial configuration. The minimum solution length for N checkers of each color is $(N + 1)^2 - 1$ moves, creating a quadratic relationship between problem size and solution complexity. This rule only applies when all the red checkers are on one side and all the blue checkers are on the other.

Our Checkers Jumping dataset generation significantly extends the approach described in \citet{shojaee2025illusion} with enhanced state space exploration, computational optimizations, and comprehensive coverage of puzzle configurations. Unlike Tower of Hanoi which has a known recursive solution, Checkers Jumping requires search to find optimal solutions. We implemented a memory-efficient Breadth-First Search (BFS) algorithm with several optimizations:

\begin{algorithm}[h]
\caption{Optimal Checkers Jumping Solver via BFS}
\label{alg:bfs}
\begin{algorithmic}[1]
\Procedure{BFS\_Solver}{$s_{start}, s_{goal}, d_{max}$}
    \State $Q \gets \text{Queue}([(s_{start}, 0)])$
    \State $\text{Visited} \gets \{s_{start}\}$
    \State $\text{Parents} \gets \{s_{start}: \text{None}\}$
    \While{$Q$ is not empty}
        \State $(s, \text{depth}) \gets Q.\text{popleft}()$
        \If{$\text{depth} \geq d_{max}$}
            \State \textbf{continue}
        \EndIf
        \For{each move $m \in \text{GetMoves}(s)$}
            \State $s' \gets \text{ApplyMove}(s, m)$
            \If{$s' \notin \text{Visited}$}
                \State $\text{Parents}[s'] \gets s$
                \If{$s' = s_{goal}$}
                    \State \Return \Call{ReconstructPath}{$\text{Parents}, s_{start}, s'$}
                \EndIf
                \State $\text{Visited}.\text{add}(s')$
                \State $Q.\text{append}((s', \text{depth} + 1))$
            \EndIf
        \EndFor
    \EndWhile
    \State \Return \text{None} \Comment{No solution found within depth limit}
\EndProcedure
\end{algorithmic}
\end{algorithm}

Instead of storing complete solution paths for each explored state (which causes exponential memory growth), we maintain only parents (dictionary mapping each state to its predecessor) and \texttt{move\_to\_state} (dictionary mapping each state to the move that created it). Path reconstruction occurs only when the goal is found, working backward from goal to start. This optimization reduces memory consumption from $O(b^d \times d)$ to $O(b^d)$ where b is the branching factor and d is the solution depth.

\begin{figure}[H]
    \centering
    \includegraphics[page=2, width=0.8\textwidth]{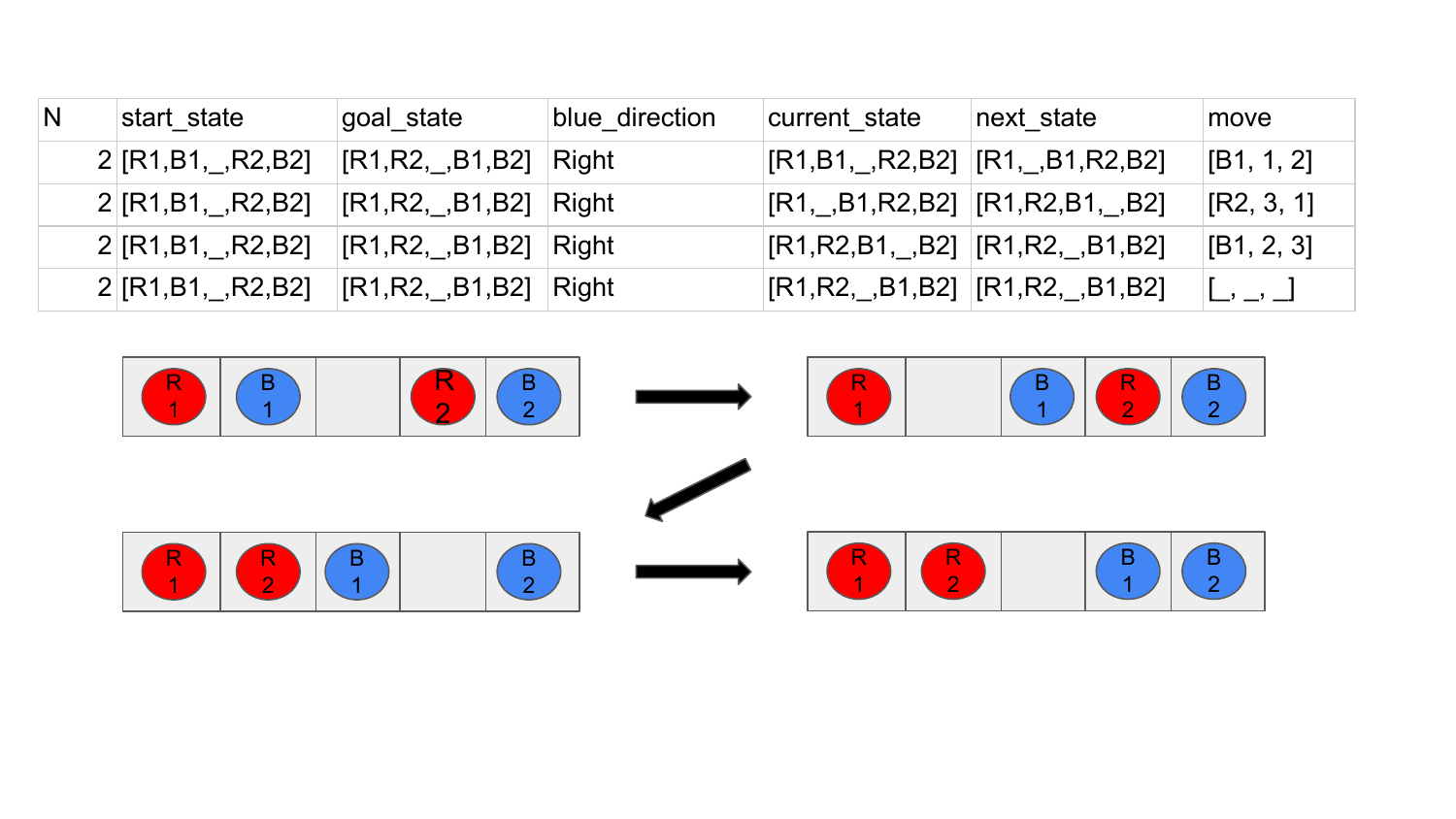}
    \caption{Example Checkers Jumping puzzle with $N=3$ checkers per colour, illustrating the initial and goal configurations.}
    \label{fig:puzzle_cj}
\end{figure}

Figure~\ref{fig:puzzle_cj} shows an example Checkers Jumping configuration.

\section{River Crossing}
River Crossing is a constraint satisfaction planning puzzle that tests multi-agent coordination and safety constraint management. This puzzle is a generalization of classic problems such as the Missionaries and Cannibals problem and the Bridge and Torch problem, which have been widely studied in planning literature. The puzzle involves N actors (denoted a1, a2, ..., an) and their corresponding N agents (denoted A1, A2, ..., An) who must cross a river using a boat with capacity k. In the initial state, all 2N individuals are on one bank of the river (typically the left bank). The goal is to transport everyone safely to the opposite bank.

The puzzle operates under three fundamental constraints. (1) Boat Capacity Constraint: The boat can carry at most k individuals at a time. (2) Non-Empty Boat Constraint: The boat cannot travel empty and must have at least one person aboard. (3) Safety Constraint: An actor ai cannot be in the presence of another agent Aj (where $j \neq i$) unless their own agent Ai is also present. This applies both on the banks and in the boat. The safety constraint creates a complex planning challenge: each agent must protect their client (actor) from competing agents. Violation of this constraint at any point (on either bank or during transit) renders the solution invalid.

The puzzle complexity can be controlled by adjusting the number of actor-agent pairs, the boat capacity, and left or right initial configuration. Our River Crossing dataset generation extends the approach described in \citet{shojaee2025illusion} with comprehensive configuration coverage, pair permutation augmentations, and memory-efficient parallel processing.

To explore the planning complexity landscape and model adaptability to resource constraints, we evaluate three boat capacity values: $k=2$ (minimal capacity for $N \in \{2, 3\}$), $k=3$ (medium capacity for $N \geq 4$), and $k=4$ (large capacity for all $N \leq 10$). This multi-capacity approach accounts for the inherent solvability limits dictated by safety constraints; specifically, our preliminary analysis confirmed that the puzzle is unsolvable for $N > 3$ when $k=2$ and for $N > 5$ when $k=3$, whereas $k=4$ remains solvable across all tested configurations.

River Crossing requires a search-based approach to identify valid solutions. We implemented a Breadth-First Search (BFS) algorithm, detailed in Algorithm \ref{alg:river_crossing}, which exhaustively explores the state space while enforcing all safety and capacity constraints.

\begin{algorithm}[h]
\caption{River Crossing Optimal Solver via BFS}
\label{alg:river_crossing}
\begin{algorithmic}[1]
\Procedure{BFSRiverCrossing}{$S_{start}, S_{goal}, B_{start}$}
    \State $Q \gets \text{Queue}([(S_{start}, B_{start}, [])])$ \Comment{Queue stores (state, boat\_side, path)}
    \State $\text{Visited} \gets \{(S_{start}, B_{start})\}$
    \While{$Q$ is not empty}
        \State $(S, B, \mathcal{P}) \gets Q.\text{popleft}()$
        \If{$S = S_{goal}$}
            \State \Return $\mathcal{P}$ \Comment{Return optimal sequence of moves}
        \EndIf
        \For{each $(S', B', m) \in \text{Successors}(S, B)$}
            \If{$(S', B') \notin \text{Visited}$}
                \State $\text{Visited}.\text{add}((S', B'))$
                \State $Q.\text{append}((S', B', \mathcal{P} \cup \{m\}))$
            \EndIf
        \EndFor
    \EndWhile
    \State \Return \text{None} \Comment{No valid solution exists}
\EndProcedure
\end{algorithmic}
\end{algorithm}

A core contribution of our dataset is the systematic generation of initial state permutations. Rather than restricting the task to the canonical starting configuration (where all entities reside on a single bank), we generate intermediate configurations where a subset of actor-agent pairs has already transitioned. This approach ensures the model learns generalizable crossing rules rather than memorizing a fixed sequence from a singular start state.

\begin{algorithm}[H]
\caption{Systematic Pair Permutation Generation}
\label{alg:pair_perm}
\begin{algorithmic}[1]
\Procedure{GeneratePermutations}{$N$}
    \State $\mathcal{X} \gets \emptyset$ \Comment{Set of all valid initial configurations}
    \For{$k = 0$ \textbf{to} $N$}
        \State $\mathcal{C} \gets \text{combinations}(\{1, \dots, N\}, k)$ \Comment{Subsets of indices of size $k$}
        \For{each index set $I \in \mathcal{C}$}
            \State $L \gets \{(a_i, A_i) \mid i \notin I\}$ \Comment{Pairs remaining on start bank}
            \State $R \gets \{(a_i, A_i) \mid i \in I\}$ \Comment{Pairs already on opposite bank}
            \State $\mathcal{X} \gets \mathcal{X} \cup \{(L, R)\}$
        \EndFor
    \EndFor
    \State \Return $\mathcal{X}$
\EndProcedure
\end{algorithmic}
\end{algorithm}

This augmentation provides several benefits. (1) Richer training signal: Models see puzzles at different stages of completion. (2) Partial solution learning: Models learn to complete puzzles from intermediate states. (3) Generalization testing: Evaluates whether models can solve non-canonical starting positions. (4) Data efficiency: Generates diverse puzzles without running full BFS from scratch for each. In addition to pair permutations, we generate puzzles starting from both banks. This tests spatial invariance: can models solve the puzzle regardless of which bank is the starting point?

States are formatted as tuples of sorted lists: (['a1', 'A1'], ['a2', 'A2', 'a3', 'A3']) representing left and right banks. Moves are lists of entities traveling together. For example,  ['a1', 'A1'] indicates actor 1 and agent 1 crossing together. The \texttt{boat\_side} column stores the current location of the boat before the move ('L' or 'R'). \texttt{goal\_direction} column denotes target bank ("Left" or "Right"). \texttt{Total\_moves} column holds the complete solution length.

\begin{figure}[H]
    \centering
    \includegraphics[page=3, width=0.8\textwidth]{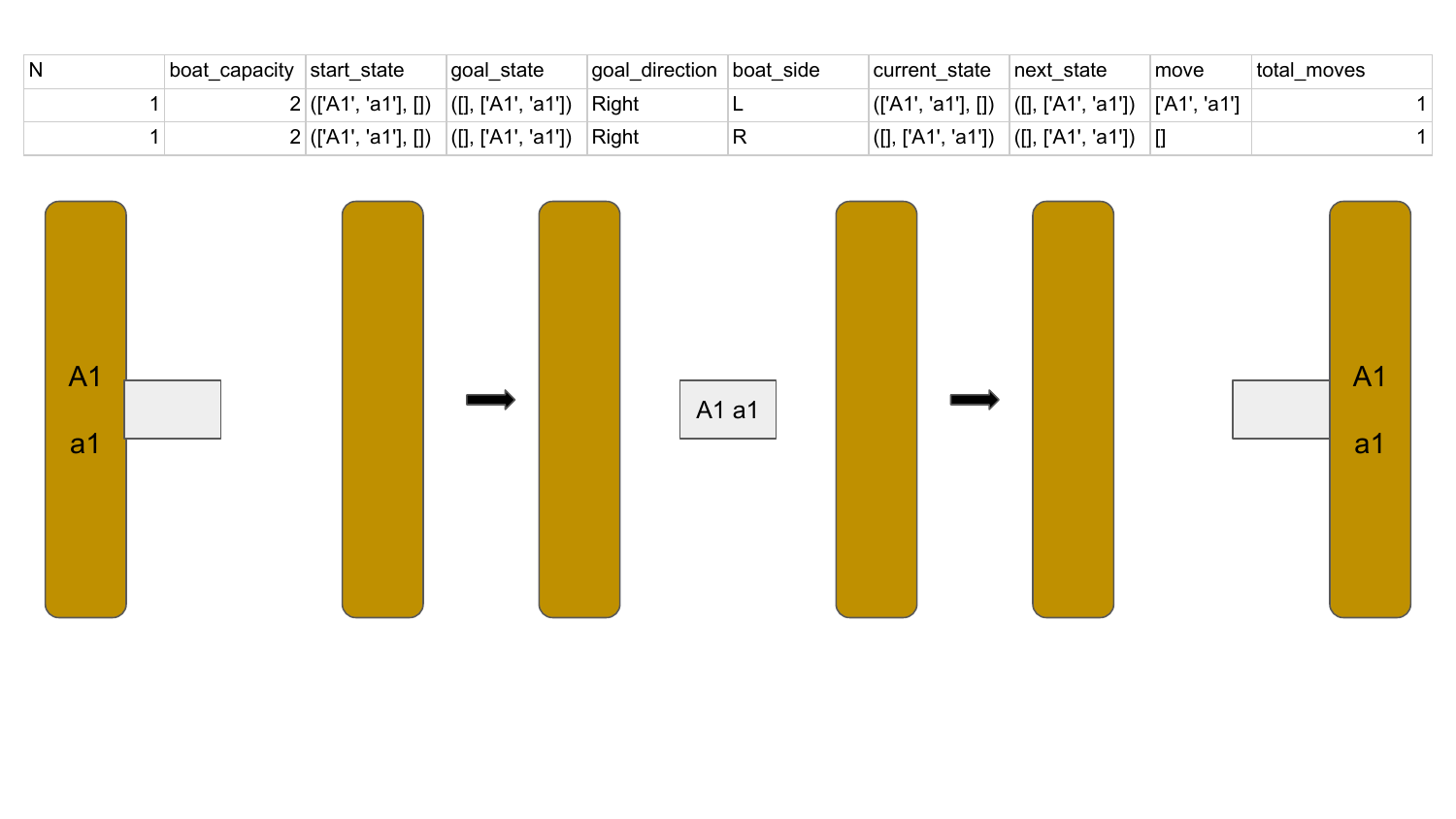}
    \caption{Example River Crossing puzzle with $N=3$ actor-agent pairs, showing the safety constraint and boat crossing mechanics.}
    \label{fig:puzzle_rc}
\end{figure}

Figure~\ref{fig:puzzle_rc} illustrates the River Crossing puzzle setup with actors and agents on the two banks.

\section{Block World}
Blocks World is a classical planning puzzle that has been extensively studied in AI planning literature and recently examined for analyzing the planning capabilities of Large Language Models. The puzzle involves multiple uniquely labeled blocks (A, B, C, etc.) arranged in stacks, where the objective is to rearrange blocks from an initial configuration to a specified goal configuration. The puzzle operates under two fundamental constraints: (1) Top Block Movement: Only the topmost block from any stack can be moved. (2) Valid Placement: A block can only be placed either on an empty stack position or on top of another block. These simple constraints create a complex planning problem where the order of operations becomes critical. Some configurations require temporary placement of blocks to access those beneath them, necessitating multi-step look ahead planning and state space exploration.

Blocks World serves as an excellent testbed for evaluating sequential planning capabilities because it requires: (1)Dependency reasoning: Understanding which blocks must be moved before others can be accessed. (2) Subgoal decomposition: Breaking complex rearrangements into achievable intermediate steps. (3) State tracking: Maintaining awareness of all block positions throughout the solution (4) Efficient path planning: Finding solutions without unnecessary moves. The puzzle difficulty can be scaled by adjusting the number of blocks and the number of stacks

Our Blocks World dataset generation significantly extends the approach described in \citet{shojaee2025illusion} with advanced memory optimizations, systematic configuration patterns, and comprehensive parallelization strategies inspired by our successful Checkers Jumping implementation. Following the successful pattern from Checkers Jumping, we implemented parent pointer tracking instead of storing complete solution paths during BFS exploration. To ensure computational feasibility across a high volume of puzzle configurations, we implement a memory-optimized Breadth-First Search (BFS) as described in Algorithm \ref{alg:bfs_optimized}. By utilizing a parent-pointer map $\pi$, we avoid the memory overhead of storing complete partial trajectories in the search queue, and we impose a node exploration limit $N_{max}$ to bound search time.

\begin{algorithm}[H]
\caption{Memory-Efficient BFS with Parent Tracking}
\label{alg:bfs_optimized}
\begin{algorithmic}[1]
\Procedure{OptimizedBFS}{$s_{start}, s_{goal}, d_{max}, N_{max}$}
    \State $Q \gets \text{Queue}([(s_{start}, 0)])$
    \State $\pi \gets \{s_{start} \mapsto \text{None}\}$ \Comment{Map of state to (parent, move)}
    \State $n_{nodes} \gets 0$
    \While{$Q$ is not empty \textbf{and} $n_{nodes} < N_{max}$}
        \State $(s, d) \gets Q.\text{popleft}()$
        \State $n_{nodes} \gets n_{nodes} + 1$
        \If{$d \geq d_{max}$} \textbf{continue} \EndIf
        \For{each move $m \in \text{GetValidMoves}(s)$}
            \State $s' \gets \text{ApplyMove}(s, m)$
            \If{$s' = s_{goal}$}
                \State \Return \Call{ReconstructPath}{$\pi, s_{start}, s, m$}
            \EndIf
            \If{$s' \notin \text{domain}(\pi)$}
                \State $\pi[s'] \gets (s, m)$
                \State $Q.\text{append}((s', d + 1))$
            \EndIf
        \EndFor
    \EndWhile
    \State \Return \text{None}
\EndProcedure
\end{algorithmic}
\end{algorithm}

We apply the same parent-pointer BFS optimization described in the Checkers Jumping section above.

To ensure a balanced and high-quality dataset despite the potential for search failures in randomly generated configurations, we employ a parallel oversampling strategy. For each target configuration $(N, K)$, we generate a task set $\mathcal{T}$ of size $M \cdot \alpha$, where $M$ is the required number of puzzles and $\alpha=3$ is the oversampling factor. These tasks are solved in parallel to extract the first $M$ valid trajectories, as formalized in Algorithm \ref{alg:oversampling}. For each N value requesting 100 puzzles, we generate 300 candidates ($3\times$ oversampling), then select the first 100 successful solutions. This ensures up to 100 puzzles per N value even if some attempts fail, we obtain sufficient puzzles

\begin{algorithm}[h]
\caption{Parallel Dataset Oversampling}
\label{alg:oversampling}
\begin{algorithmic}[1]
\Procedure{GeneratePuzzles}{$N, K, M, \alpha, P$}
    \State $\mathcal{T} \gets \{(N, K, \text{seed}_i) \mid i = 1, \dots, M \cdot \alpha\}$ \Comment{Create oversampled task set}
    \State $\mathcal{R} \gets \text{ParallelMap}(\text{SolvePuzzle}, \mathcal{T}, \text{workers}=P)$
    \State $\mathcal{D} \gets \{r \in \mathcal{R} \mid r \neq \text{None}\}$ \Comment{Filter successful search results}
    \If{$|\mathcal{D}| < M$}
        \State \textbf{warn} "Insufficient valid puzzles generated"
    \EndIf
    \State \Return $\mathcal{D}[1 \dots M]$ \Comment{Truncate to required dataset size}
\EndProcedure
\end{algorithmic}
\end{algorithm}

\begin{figure}[h]
    \centering
    \includegraphics[page=4,width=0.8\textwidth]{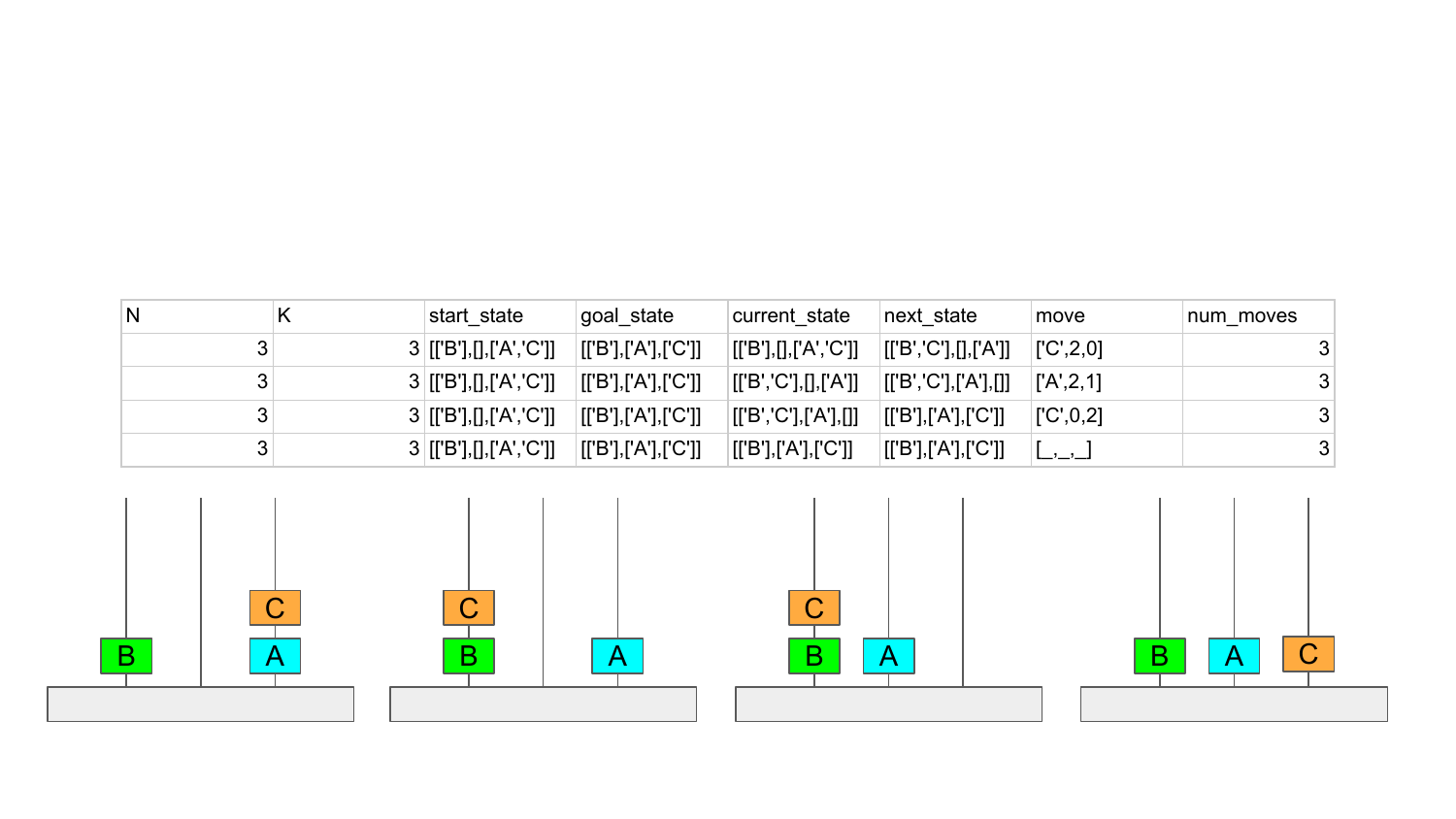}
    \caption{Example Block World puzzle with $N=4$ blocks across 3 stacks, showing the initial and goal configurations.}
    \label{fig:puzzle_bw}
\end{figure}

Figure~\ref{fig:puzzle_bw} shows an example Block World rearrangement problem.

\section{Notation and Definitions}
\label{app:notation}

Table~\ref{tab:notation} defines all mathematical notation used in
Sections~\ref{sec:results} and~\ref{sec:analysis}.

\begin{table}[h]
  \caption{Summary of notation used throughout the paper.}
  \label{tab:notation}
  \centering
  \setlength{\tabcolsep}{8pt}
  \begin{tabular}{lp{0.70\textwidth}}
    \toprule
    \textbf{Symbol} & \textbf{Definition} \\
    \midrule
    $S$ & Finite set of all possible puzzle states. \\
    $s_0$ & Initial state of a puzzle instance. \\
    $s_g$ & Goal state; a solution must reach $s_g$ from $s_0$. \\
    $S_{\text{valid}} \subseteq S$ & Subset of states satisfying all hard puzzle
      constraints (e.g.\ no larger disk on a smaller one in ToH; safety rule
      satisfied on every bank in RC; only top-of-stack blocks moved in BW). \\
    $\hat{s}_t$ & Predicted state at rollout step $t$; $\hat{s}_0 = s_0$. \\
    $\pi_\theta$ & Learned transition function parameterised by $\theta$;
      $\hat{s}_{t+1} = \pi_\theta(\hat{s}_t)$. \\
    $T^*$ & The rollout step at which the goal is first reached,
      i.e.\ the smallest $t$ with $\hat{s}_t = s_g$. \\
    $T_{\max}$ & Maximum allowed rollout steps; set per instance to
      $2 \times |\tau^\star|$ (twice the BFS-optimal solution length). \\
    $\tau^\star$ & BFS-optimal trajectory for a given instance;
      $|\tau^\star|$ is its length in moves. \\
    $\hat{\tau}$ & Model rollout trajectory; $|\hat{\tau}|$ is its length. \\
    $L(N)$ & Expected optimal solution length as a function of difficulty $N$. \\
    $\varepsilon$ & Per-step error rate in the compounding error model
      (Eq.~\eqref{eq:compounding}). \\
    \midrule
    \multicolumn{2}{l}{\textit{Failure modes (assigned at first failure event):}} \\
    \textbf{invalid\_move} & $\hat{s}_{t+1} \notin S_{\text{valid}}$; a puzzle
      constraint is violated. \\
    \textbf{invalid\_output} & $\hat{s}_{t+1}$ cannot be parsed into the puzzle's
      state grammar. \\
    \textbf{loop} & $\hat{s}_{t+1} = \hat{s}_j$ for some $j < t$; a previously
      visited state is revisited. \\
    \textbf{premature\_stop} & $\langle\texttt{STOP}\rangle$ emitted at step $t$
      with $\hat{s}_t \neq s_g$; the model halts before reaching the goal. \\
    \bottomrule
  \end{tabular}
\end{table}

\section{Full Dataset Statistics}
\label{app:dataset}

\begin{table}[h]
  \caption{Full RecurrReason benchmark statistics.}
  \centering
  \resizebox{\textwidth}{!}{%
  \begin{tabular}{l rr rr rr rr}
    \toprule
    & \multicolumn{2}{c}{\textbf{Block World}}
    & \multicolumn{2}{c}{\textbf{Checkers Jumping}}
    & \multicolumn{2}{c}{\textbf{Tower of Hanoi}}
    & \multicolumn{2}{c}{\textbf{River Crossing}} \\
    \cmidrule(lr){2-3}\cmidrule(lr){4-5}\cmidrule(lr){6-7}\cmidrule(lr){8-9}
    $N$ & \# Puzzles & Moves & \# Puzzles & Moves
        & \# Puzzles & Moves & \# Puzzles & Moves \\
    \midrule
    1  & 6   & 6      & 4    & 6       & 6  & 6      & 6    & 6      \\
    2  & 52  & 105    & 12   & 52      & 6  & 18     & 18   & 30     \\
    3  & 92  & 336    & 40   & 336     & 6  & 42     & 42   & 110    \\
    4  & 99  & 497    & 140  & 1{,}928   & 6  & 90     & 60   & 156    \\
    5  & 100 & 615    & 504  & 10{,}260  & 6  & 186    & 124  & 456    \\
    6  & 100 & 750    & 1000 & 28{,}106  & 6  & 378    & 126  & 414    \\
    7  & 100 & 905    & 1000 & 35{,}924  & 6  & 762    & 254  & 1{,}058  \\
    8  & 100 & 915    & 1000 & 45{,}076  & 6  & 1{,}530  & 510  & 2{,}598  \\
    9  & 100 & 861    & 1000 & 54{,}912  & 6  & 3{,}066  & 1022 & 6{,}186  \\
    10 & 100 & 837    & 1000 & 65{,}894  & 6  & 6{,}138  & 2046 & 14{,}382 \\
    \midrule
    \textbf{Total} & 849 & 5{,}827 & 5700 & 242{,}494
                   & 60 & 12{,}216 & 4208 & 25{,}396 \\
    \bottomrule
  \end{tabular}%
  }
\end{table}

\section{Full Per-Puzzle Result Tables}
\label{app:full_tables}

\begin{table}[H]
  \caption{Checkers Jumping: full results.}
  \label{tab:cj_baselines}
  \centering
  \begin{tabular}{llcc}
    \toprule
    Model & Condition & Val (\%) & OOD (\%) \\
    \midrule
    T5 (scratch)         & Trained   & 0.00 & 0.00 \\
    T5 (pre-trained)     & Zero-shot & 0.00 & 0.00 \\
    T5 (pre-trained)     & Fine-tuned& 1.11 & 0.10 \\
    \midrule
    GPT-2 (scratch)      & Trained   & 1.11 & 0.03 \\
    GPT-2 (pre-trained)  & Zero-shot & 0.00 & 0.00 \\
    GPT-2 (pre-trained)  & Fine-tuned& 1.11 & 0.10 \\
    \bottomrule
  \end{tabular}
  \begin{minipage}{\linewidth}
    \vspace{4pt}

  \end{minipage}
\end{table}

\begin{table}[H]
  \caption{Tower of Hanoi: full results.}
  \label{tab:toh_results}
  \centering
  \begin{tabular}{llcc}
    \toprule
    Model & Condition & Val (\%) & OOD (\%) \\
    \midrule
    T5 (scratch)         & Trained   & 0.00  & 0.00 \\
    T5 (pre-trained)     & Zero-shot & 0.00  & 0.00 \\
    T5 (pre-trained)     & Fine-tuned& 11.11 & 0.00 \\
    GPT-2 (scratch)      & Trained   & 0.00  & 0.00 \\
    GPT-2 (pre-trained)  & Zero-shot & 0.00  & 0.00 \\
    GPT-2 (pre-trained)  & Fine-tuned& 0.00  & 0.00 \\
    \bottomrule
  \end{tabular}
\end{table}

\begin{table}[H]
  \caption{River Crossing: full results. All entries are $0.00\%$.}
  \label{tab:rc_results}
  \centering
  \begin{tabular}{llcc}
    \toprule
    Model & Condition & Val (\%) & OOD (\%) \\
    \midrule
    T5 (scratch)         & Trained   & 0.00 & 0.00 \\
    T5 (pre-trained)     & Zero-shot & 0.00 & 0.00 \\
    T5 (pre-trained)     & Fine-tuned& 0.00 & 0.00 \\
    GPT-2 (scratch)      & Trained   & 0.00 & 0.00 \\
    GPT-2 (pre-trained)  & Zero-shot & 0.00 & 0.00 \\
    GPT-2 (pre-trained)  & Fine-tuned& 0.00 & 0.00 \\
    \bottomrule
  \end{tabular}
\end{table}

\begin{table}[H]
  \caption{Block World: full results.}
  \label{tab:bw_results}
  \centering
  \begin{tabular}{llcc}
    \toprule
    \textbf{Model} & \textbf{Condition} & \textbf{Val (\%)} & \textbf{OOD (\%)} \\
    \midrule
    T5 (scratch)         & Trained   & 0.00           & 0.00 \\
    T5 (pre-trained)     & ZS        & 0.00           & 0.00 \\
    T5 (pre-trained)     & FT        & 97.27 & 81.00 \\
    GPT-2 (scratch)      & Trained   & 21.82          & 0.00 \\
    GPT-2 (pre-trained)  & ZS        & 0.00           & 0.00 \\
    GPT-2 (pre-trained)  & FT        & 24.55          & 0.00 \\
    \bottomrule
  \end{tabular}
\end{table}
\section{Per-Puzzle Failure Mode Breakdowns}
\label{app:failure_tables}

\begin{table}[H]
  \caption{Failure mode breakdown (validation, fine-tuned/trained conditions).
           Values are percentages of failed rollouts only.}
  \centering
  \setlength{\tabcolsep}{6pt}
  \begin{tabular}{llrrrr}
    \toprule
    \textbf{Puzzle} & \textbf{Model}
      & \textbf{Inv.\ Move (\%)} & \textbf{Inv.\ Output (\%)}
      & \textbf{Loop (\%)} & \textbf{Prem.\ Stop (\%)} \\
    \midrule
    \multirow{3}{*}{Block World}
      & T5 PT FT (Val)      &   0.0 &  0.0 & 66.7 & 33.3 \\
      & T5 PT FT (OOD)      &   4.2 &  0.0 & 58.3 & 37.5 \\
      & GPT-2 PT FT (Val)   &   0.0 & 12.0 & 80.7 &  7.2 \\
      & GPT-2 PT FT (OOD)   &  91.4 &  5.0 &  3.6 &  0.0 \\
    \cmidrule(l){2-6}
    \multirow{2}{*}{Tower of Hanoi}
      & T5 PT FT (Val)      & 100.0 &  0.0 &  0.0 &  0.0 \\
      & GPT-2 PT FT (Val)   &  33.3 &  0.0 &  0.0 & 66.7 \\
    \cmidrule(l){2-6}
    \multirow{2}{*}{Checkers Jumping}
      & T5 PT FT (Val)      &  86.5 &  0.0 & 13.1 &  0.4 \\
      & GPT-2 PT FT (Val)   &  87.3 &  3.1 &  8.4 &  1.2 \\
    \cmidrule(l){2-6}
    \multirow{2}{*}{River Crossing}
      & T5 PT FT (Val)      &  62.7 &  0.0 &  0.0 & 37.3 \\
      & T5 PT FT (OOD)      &  92.4 &  0.0 &  0.0 &  7.6 \\
    \bottomrule
  \end{tabular}
\end{table}

\section{Model Hyperparameters}
\label{app:hparams}

\begin{table}[H]
  \caption{Training hyperparameters for all model families.}
  \centering
  \begin{tabular}{lcc}
    \toprule
    & \textbf{T5} & \textbf{GPT-2} \\
    \midrule
    Parameters     & 60.5M & 124M      \\
    Optimizer      & AdamW & AdamW     \\
    Learning rate  & $10^{-4}$ & $10^{-4}$ \\
    Batch size     & 16    & 16        \\
    Stopping       & patience=5 & patience=5 \\
    Pre-training   & C4    & WebText   \\
    \bottomrule
  \end{tabular}
\end{table}

\section{Learnability Analysis: Derivations and Complexity}
\label{app:learnability}

Section~5 of the main text argues that three structural properties of a
puzzle determine its learnability ceiling for autoregressive sequence
models: \emph{transition locality}, \emph{action space size}, and
\emph{solution length growth}.  This appendix provides formal
derivations for each quantity and spells out the compounding-error model
that ties them together.

\subsection{Solution Length Derivations}
\label{app:sol_length}

For every puzzle we derive the optimal (BFS-verified) solution length
$L(N)$ as a function of the difficulty parameter~$N$.

\paragraph{Block World: $L(N) = O(N)$.}

\paragraph{Setup.}
$N$ uniquely labelled blocks are distributed across $K$ stacks.
A single move takes the top block of one stack and places it on another
stack (or on an empty stack position).

\paragraph{Worst-case argument.}
In the worst case every block must be moved exactly once: the initial
and goal configurations share no block in its correct final position.
Because each move repositions exactly one block, at least $N$ moves are
required.  Conversely, when $K \geq 3$ there is always at least one
empty stack (or a stack whose top block is already in its goal position)
that can serve as a temporary buffer, so $N$ moves are also sufficient
in the worst case with enough stacks.  More precisely, when stacks are
limited, some blocks may need to be moved aside temporarily and then
returned, but the total number of moves remains bounded linearly:

\begin{equation}
  L_{\mathrm{BW}}(N) = \Theta(N).
  \label{eq:l_bw}
\end{equation}

\paragraph{Checkers Jumping: $L(N) = (N+1)^{2} - 1$.}

\paragraph{Setup.}
A linear board of $2N+1$ cells contains $N$ red checkers on the left,
one gap in the centre, and $N$ blue checkers on the right.  Red may only
move right; blue may only move left.  A checker may \emph{slide} one
cell forward into the gap or \emph{jump} over exactly one checker of the
opposite colour into the gap.  The goal is to swap the red and blue
groups.

\paragraph{Derivation.}
The optimal strategy alternates slides and jumps in a structured
pattern.  Each of the $N$ red checkers must cross all $N$ blue
checkers (one jump each) and also traverse the gap (one slide).
Symmetrically, each blue checker does the same.  The total number of
jumps is $N \times N = N^2$ (each red jumps over each blue once) and
the total number of slides is $N + N = 2N$ (each checker slides
exactly once).  This gives:
\begin{equation}
  L_{\mathrm{CJ}}(N)
  = N^2 + 2N
  = (N+1)^{2} - 1.
  \label{eq:l_cj}
\end{equation}

\paragraph{Tower of Hanoi: $L(N) = 2^{N} - 1$.}

\paragraph{Setup.}
$N$ disks of distinct sizes sit on 3 pegs.  Only the top disk of a peg
may be moved, and a larger disk may never rest on a smaller one.  The
goal is to transfer all disks from the source peg to the target peg.

\paragraph{Derivation}
Let $T(n)$ denote the minimum number of moves for $n$ disks.  The
base case is $T(1) = 1$.  For $n > 1$:
\begin{enumerate}[leftmargin=2em]
  \item Move the top $n-1$ disks from source to auxiliary peg: $T(n-1)$
        moves.
  \item Move the largest disk from source to target: $1$ move.
  \item Move the $n-1$ disks from auxiliary to target: $T(n-1)$ moves.
\end{enumerate}
Hence
\begin{equation}
  T(n) = 2\,T(n-1) + 1, \quad T(1) = 1.
  \label{eq:toh_recurrence}
\end{equation}
Solving by substitution ($T(n) = 2^{n} - 1$ satisfies
$2(2^{n-1}-1)+1 = 2^{n}-1$) or by the change of variable
$U(n) = T(n)+1$ which gives $U(n)=2U(n-1)$, we obtain:
\begin{equation}
  L_{\mathrm{ToH}}(N) = T(N) = 2^{N} - 1.
  \label{eq:l_toh}
\end{equation}

\paragraph{Optimality.}
The Frame-Stewart lower bound for 3 pegs shows that any algorithm must
move the largest disk, and to do so the $n-1$ smaller disks must be
cleared from above it and later restacked.  Hence
$T(n) \geq 2\,T(n-1)+1$, matching the recursion above, so $2^N - 1$ is
both achievable and minimal.

\paragraph{River Crossing: $L(N, k)$ depends on $N$ and boat capacity $k$.}

\paragraph{Setup.}
$N$ actor-agent pairs must cross a river using a boat of capacity $k$.
The safety constraint requires that actor $a_i$ is never in the
presence of agent $A_j$ ($j \neq i$) unless $A_i$ is also present, on
either bank or in the boat.

\paragraph{Solution length.}
Unlike the previous puzzles, River Crossing has no single closed-form
$L(N)$ because the solution length depends jointly on $N$ and $k$.
The first forward trip carries up to $k$ individuals.  Each subsequent
round-trip (one return, one forward) moves a net of at most $k-1$
individuals across, since at least one person must return with the
boat.  After the first trip transports $k$ people, the remaining
$2N - k$ require $\lceil(2N-k)/(k-1)\rceil$ round-trips, each
consisting of 2 crossings.  Including the initial forward trip, a
transport-only lower bound is:
\begin{equation}
  L_{\mathrm{RC}}(N, k) \;\geq\; 2\!\left\lceil \frac{2N - k}{k-1} \right\rceil + 1.
  \label{eq:l_rc_lower}
\end{equation}
For $N{=}2, k{=}2$ this gives $2\lceil 2/1\rceil + 1 = 5$, matching
the BFS optimum.  However, the safety constraint often forces
suboptimal groupings, so the actual BFS-optimal length can exceed this
transport-only lower bound.  



\subsection{Transition Locality Analysis}
\label{app:transition}

We define the \emph{transition locality} of a puzzle as the number of
tokens (or state components) that a model must inspect to verify whether
a proposed action is legal in the current state.  This directly affects
how many attention steps are needed for the model to perform correct
next-step prediction.

\paragraph{Block World: $O(1)$ locality.}

A move ``take the top block of stack $i$ and place it on stack $j$''
requires checking exactly two things:
\begin{enumerate}[leftmargin=2em]
  \item Stack $i$ is non-empty (the block to be moved exists).
  \item Stack $j$ either exists and can receive a block, or is an empty
        position.
\end{enumerate}
Both checks involve reading only the \emph{top element} of two stacks,
independent of how many blocks are below.  The number of tokens
inspected is $O(1)$ regardless of $N$.

\begin{equation}
  C_{\mathrm{verify}}^{\mathrm{BW}} = O(1).
  \label{eq:trans_bw}
\end{equation}

\paragraph{Checkers Jumping: $O(N)$ locality.}

A checker at position $p$ may slide to $p \pm 1$ (if empty) or jump to
$p \pm 2$ (if $p \pm 1$ holds an opposite-colour checker and $p \pm 2$
is empty).  While each individual move check is $O(1)$ in the board
representation, the directional constraint (red moves only right, blue
moves only left) requires knowing the \emph{colour} of the checker at
position $p$, which in our serialized representation requires the model
to parse the full board state to determine which checkers are which.
Furthermore, to determine whether a given move leads to a dead-end
(a configuration from which no solution exists), the model must
effectively evaluate the positions of all $2N$ checkers.  Hence:
\begin{equation}
  C_{\mathrm{verify}}^{\mathrm{CJ}} = O(N).
  \label{eq:trans_cj}
\end{equation}

\paragraph{Tower of Hanoi: $O(N)$ locality.}

Moving disk $d$ from peg $i$ to peg $j$ requires verifying:
\begin{enumerate}[leftmargin=2em]
  \item Disk $d$ is on top of peg $i$ (no smaller disk above it).
  \item No disk on peg $j$ is smaller than $d$.
\end{enumerate}
Check (2) requires comparing $d$ with the top disk on peg $j$.  In the
worst case, up to $N-1$ disks may be on peg $j$, and the model must
parse the peg contents to find the top disk.  In our serialized
state representation (a list of three lists), identifying the top disk
of a peg requires scanning through the state tokens.  Since the total
state has $\Theta(N)$ tokens distributed across 3 pegs, locating and
comparing the relevant disk sizes requires:
\begin{equation}
  C_{\mathrm{verify}}^{\mathrm{ToH}} = O(N).
  \label{eq:trans_toh}
\end{equation}

\paragraph{River Crossing: $O(N)$ global constraint.}

After every boat trip, the safety constraint must be verified on
\emph{both} banks and inside the boat.  For a single bank with
entities $E$, checking whether actor $a_i \in E$ is safe requires
verifying that for every agent $A_j \in E$ with $j \neq i$, the
corresponding agent $A_i$ is also in $E$.  In the worst case all $2N$
entities are distributed across the two banks, so:
\begin{equation}
  C_{\mathrm{verify}}^{\mathrm{RC}} = O(N)
  \quad \text{(per entity, summing to } O(N^2) \text{ for the full
  state)}.
  \label{eq:trans_rc}
\end{equation}
This is a \emph{global} constraint: every entity's safety depends on
which other entities share its location.  A model must attend to the
full state representation at each step, in contrast to Block World's
purely local top-of-stack check.

\subsection{Action Space Analysis}
\label{app:action_space}

The \emph{branching factor} $b$ measures the number of legal actions
available in a typical state.  Larger branching factors make
next-step prediction harder because the model must distinguish the
correct action from more alternatives.

\paragraph{Block World: $O(K^{2})$.}

With $K$ stacks, a move is specified by a source stack and a
destination stack ($i \neq j$).  The number of possible moves is
at most $K(K-1)$.  In practice, only non-empty source stacks contribute,
but the upper bound is:
\begin{equation}
  b_{\mathrm{BW}} = O(K^2).
  \label{eq:action_bw}
\end{equation}
In our benchmark $K = 3$, giving at most $3 \times 2 = 6$ moves per
state.  This is a small, fixed branching factor independent of $N$.

\paragraph{Checkers Jumping: $O(1)$.}

Only checkers adjacent to the single gap can move: at most two cells
at slide distance ($p \pm 1$) and two at jump distance ($p \pm 2$),
giving at most 4 candidate moves per state regardless of $N$.
Directional constraints further reduce this:
\begin{equation}
  b_{\mathrm{CJ}} = O(1).
  \label{eq:action_cj}
\end{equation}
Empirically, the number of legal moves per state is between 1 and 3.
The difficulty of Checkers Jumping comes not from branching but from
quadratic solution depth and the need to avoid dead-end configurations.

\paragraph{Tower of Hanoi: $O(1)$.}

With 3 pegs, each non-empty peg can potentially move its top disk to
one of the other 2 pegs.  The number of legal moves is therefore at
most $3 \times 2 = 6$, but the size ordering constraint typically
reduces this.  Crucially, the branching factor is \emph{constant}:
\begin{equation}
  b_{\mathrm{ToH}} = O(1).
  \label{eq:action_toh}
\end{equation}
The difficulty of Tower of Hanoi comes not from action space breadth but
from exponential solution depth.

\paragraph{River Crossing: combinatorial $O\!\left(\binom{2N}{k}\right)$.}

On the boat-side bank, any subset of 1 to $k$ entities can be loaded
onto the boat.  The number of candidate boat loadings is:
\begin{equation}
  b_{\mathrm{RC}} = \sum_{j=1}^{k} \binom{|\text{bank}|}{j}
  \;\leq\; \sum_{j=1}^{k} \binom{2N}{j}.
  \label{eq:action_rc}
\end{equation}
For $k = 2$ and $2N$ entities on one bank, this is
$\binom{2N}{1} + \binom{2N}{2} = 2N + N(2N-1) = O(N^2)$.  For larger
$k$ the growth is polynomial in $N$ of degree $k$, i.e., $O(N^k)$.
Many of these candidates violate the safety constraint and are pruned,
but the model must still implicitly evaluate them to select a legal move.
This combinatorial branching factor, combined with the $O(N)$ global
constraint check per candidate, makes River Crossing the hardest puzzle
for autoregressive models.

\subsection{Compounding Error Model}
\label{app:compounding}

We formalise the compounding-error argument from Eq.~\eqref{eq:compounding}
in the main text.

\paragraph{Setup and independence assumption.}

Let a model produce a trajectory
$\hat{s}_0, \hat{s}_1, \ldots, \hat{s}_{L}$ of length $L = L(N)$.
We assume:
\begin{itemize}[leftmargin=2em]
  \item \textbf{Step independence.} At each step $t$, the model
        produces the correct next state with probability $1-\varepsilon$,
        independently of previous steps.  This is a simplifying
        assumption; in practice, errors may be correlated.  However, the
        assumption yields a useful \emph{upper bound} on success
        probability because positive correlations (an error at step $t$
        making step $t{+}1$ harder) can only reduce the actual success
        rate.
  \item \textbf{Uniform per-step error.}  The error rate $\varepsilon$
        is the same at every step.  In reality, $\varepsilon$ may
        increase with $t$ as the model drifts from trained
        distributions, so the uniform assumption is again optimistic.
\end{itemize}

\paragraph{Exact formula and Taylor approximation.}

Under these assumptions, the probability of producing a fully correct
trajectory of length $L$ is:
\begin{equation}
  P(\text{success}) = \prod_{t=1}^{L} (1-\varepsilon) = (1-\varepsilon)^{L}.
  \label{eq:compound_exact}
\end{equation}
Taking logarithms:
\[
  \ln P(\text{success}) = L \ln(1-\varepsilon).
\]
For small $\varepsilon$ we use the Taylor expansion
$\ln(1-\varepsilon) = -\varepsilon - \frac{\varepsilon^2}{2} - \cdots
\approx -\varepsilon$, giving:
\begin{equation}
  P(\text{success})
  \;\approx\; e^{-\varepsilon L}
  \;\approx\; 1 - \varepsilon L
  \quad (\text{when } \varepsilon L \ll 1).
  \label{eq:compound_approx}
\end{equation}
The first approximation ($e^{-\varepsilon L}$) is accurate whenever
$\varepsilon^2 L \ll 1$; the second (linear) approximation holds when
$\varepsilon L$ itself is small.

\paragraph{Per-puzzle scaling.}

Substituting each puzzle's $L(N)$:

\begin{align}
  P_{\mathrm{BW}}(N) &\approx e^{-\varepsilon \cdot c\,N}
    &\quad &\text{(linear decay in } N\text{)},
    \label{eq:p_bw} \\[4pt]
  P_{\mathrm{CJ}}(N) &\approx e^{-\varepsilon\,[(N+1)^2 - 1]}
    &\quad &\text{(quadratic decay in } N\text{)},
    \label{eq:p_cj} \\[4pt]
  P_{\mathrm{ToH}}(N) &\approx e^{-\varepsilon\,(2^N - 1)}
    &\quad &\text{(exponential decay in } N\text{)},
    \label{eq:p_toh} \\[4pt]
  P_{\mathrm{RC}}(N,k) &\approx e^{-\varepsilon \cdot L_{\mathrm{RC}}(N,k)}
    &\quad &\text{(at least linear decay in } N\text{)}.
    \label{eq:p_rc}
\end{align}

The qualitative ordering is clear: Block World's success probability
decays slowest (linearly in $N$), followed by Checkers Jumping
(quadratic), River Crossing (at least linear, compounded by the
$O(N)$ constraint verification difficulty), and Tower of Hanoi
(exponential).

\paragraph{Numerical examples.}

Table~\ref{tab:compound_examples} illustrates the compounding-error
model with $\varepsilon = 0.05$ (5\% per-step error rate).

\begin{table}[ht]
  \caption{Predicted success probability $P(\text{success}) = (1-\varepsilon)^{L(N)}$
           for $\varepsilon = 0.05$.  BW uses $L(N) = N$ (best case);
           CJ uses $L(N) = (N{+}1)^2 - 1$; ToH uses $L(N) = 2^N - 1$.
           RC values use representative BFS-optimal lengths from our
           dataset.}
  \label{tab:compound_examples}
  \centering
  \begin{tabular}{c rrr rrr rrr}
    \toprule
    & \multicolumn{3}{c}{\textbf{Block World}}
    & \multicolumn{3}{c}{\textbf{Checkers Jumping}}
    & \multicolumn{3}{c}{\textbf{Tower of Hanoi}} \\
    \cmidrule(lr){2-4}\cmidrule(lr){5-7}\cmidrule(lr){8-10}
    $N$ & $L$ & $P$ & & $L$ & $P$ & & $L$ & $P$ & \\
    \midrule
    1  & 1   & 0.950 & & 3    & 0.857 & & 1    & 0.950 & \\
    2  & 2   & 0.903 & & 8    & 0.663 & & 3    & 0.857 & \\
    3  & 3   & 0.857 & & 15   & 0.463 & & 7    & 0.698 & \\
    5  & 5   & 0.774 & & 35   & 0.166 & & 31   & 0.204 & \\
    7  & 7   & 0.698 & & 63   & 0.039 & & 127  & 0.001 & \\
    10 & 10  & 0.599 & & 120  & 0.002 & & 1023 & ${\approx}\,0$ & \\
    \bottomrule
  \end{tabular}
\end{table}

\paragraph{Interpretation.}
Even with a modest 5\% per-step error:
\begin{itemize}[leftmargin=2em]
  \item \textbf{Block World} retains $\approx 60\%$ success at
        $N{=}10$, consistent with our empirical 75\% (the model's
        actual $\varepsilon$ is lower than 5\% on BW).
  \item \textbf{Checkers Jumping} drops below 5\% by $N{=}7$ and
        is effectively zero by $N{=}10$.  Our empirical 0 to 1\%
        success rate is consistent.
  \item \textbf{Tower of Hanoi} is effectively unsolvable for $N \geq 5$
        under any non-negligible $\varepsilon$.  With 1{,}023 steps at
        $N{=}10$, even $\varepsilon = 0.001$ gives
        $P \approx e^{-1.023} \approx 0.36$.
        Our empirical observation (11\% at $N{=}1$ for T5,
        0\% everywhere else) is consistent.
  \item \textbf{River Crossing} combines moderate solution length with
        the hardest per-step constraint ($O(N)$ global verification),
        so $\varepsilon$ is itself large, driving
        $P(\text{success})$ to zero across all tested configurations.
\end{itemize}

\subsection{Summary of Complexity Properties}
\label{app:summary_table}

Table~\ref{tab:learnability_summary} consolidates the three structural
properties and their effect on learnability.

\begin{table}[ht]
  \caption{Structural complexity of each puzzle and predicted
           learnability.  $L(N)$: optimal solution length.
           $C_{\mathrm{verify}}$: tokens inspected to verify one
           transition.  $b$: branching factor.  The rightmost column
           shows the dominant factor in the compounding-error decay of
           $P(\text{success})$.}
  \label{tab:learnability_summary}
  \centering
  \begin{tabular}{lcccc}
    \toprule
    \textbf{Puzzle} & $L(N)$ & $C_{\mathrm{verify}}$
      & Branching $b$ & $P(\text{success})$ decay \\
    \midrule
    Block World
      & $O(N)$ & $O(1)$ & $O(K^2)$
      & $e^{-\varepsilon N}$ (linear) \\[2pt]
    Checkers Jumping
      & $(N{+}1)^2{-}1$ & $O(N)$ & $O(1)$
      & $e^{-\varepsilon N^2}$ (quadratic) \\[2pt]
    Tower of Hanoi
      & $2^N - 1$ & $O(N)$ & $O(1)$
      & $e^{-\varepsilon\, 2^N}$ (exponential) \\[2pt]
    River Crossing
      & $\geq \Omega(N/k)$ & $O(N)$ & $O(N^k)$
      & $e^{-\varepsilon\, L(N,k)}$ (at least linear) \\
    \bottomrule
  \end{tabular}
\end{table}

\end{document}